\documentclass[runningheads]{llncs}
\usepackage{graphicx}
\usepackage{amsmath,amssymb} 
\usepackage{color}
\usepackage[width=122mm,left=12mm,paperwidth=146mm,height=193mm,top=12mm,paperheight=217mm]{geometry}

\usepackage{times}
\usepackage{epsfig}
\usepackage{graphicx}
\usepackage{amsmath}
\usepackage{amssymb}
\usepackage{authblk}
\usepackage{xspace}
\usepackage{float}
\usepackage{multirow}
\usepackage{pbox}
\usepackage{tabularx}
\newcommand{\tx}{\textit}

\def\fig#1{fig.\,\ref{#1}}

\usepackage{comment}

\newcommand{\N}{\mathcal{N}} 
\newcommand{\I}{\mathcal{I}} 
\newcommand{\D}{\mathcal{D}} 

\newcommand{\mbx}{\mathbf} %

\dblfloatsep\floatsep
\dbltextfloatsep\textfloatsep
\makeatletter
\renewcommand\normalsize{%
   \@setfontsize\normalsize\@xpt\@xiipt
   \abovedisplayskip 6\p@ \@plus2\p@ \@minus3\p@
   \abovedisplayshortskip \z@ \@plus3\p@
   \belowdisplayshortskip 4\p@ \@plus3\p@ \@minus3\p@
   \belowdisplayskip \abovedisplayskip
   \let\@listi\@listI}
\normalsize

\begin{document}
\pagestyle{headings}
\mainmatter
\def\ECCV14SubNumber{1525}  

\title{Constrained Parametric Proposals and Pooling Methods\\
for Semantic Segmentation in RGB-D Images}

\titlerunning{Semantic Segmentation in RGB-D Images}

\authorrunning{D. Banica, C. Sminchisescu}

\author{
	Dan Banica$^1$, Cristian Sminchisescu$^{2,1}$
}
\institute{	$^1$Institute of Mathematics of the Romanian Academy,
	$^2$Lund University
	{\\ \tt{\scriptsize{dan.banica@imar.ro, cristian.sminchisescu@math.lth.se}}}
}

\maketitle

\vspace{-5mm}
\begin{abstract}
We focus on the problem of semantic segmentation based on RGB-D data, with emphasis on analyzing cluttered indoor scenes containing many instances from many visual categories. Our approach is based on a parametric figure-ground intensity and depth-constrained proposal process that generates spatial layout hypotheses at multiple locations and scales in the image followed by a sequential inference algorithm that integrates the proposals into a complete scene estimate. Our contributions can be summarized as proposing the following: (1) a generalization of parametric max flow figure-ground proposal methodology to take advantage of intensity and depth information, in order to systematically and efficiently generate the breakpoints of an underlying spatial model in polynomial time, (2) new region description methods based on second-order pooling over multiple features constructed using both intensity and depth channels, (3) an inference procedure that can resolve conflicts in overlapping spatial partitions, and handles scenes with a large number of objects category instances, of very different scales, (4) extensive evaluation of the impact of depth, as well as the effectiveness of a large number of descriptors, both pre-designed and automatically obtained using deep learning, in a difficult RGB-D semantic segmentation problem with 92 classes. We report state of the art results in the challenging NYU Depth v2 dataset \cite{Silberman:ECCV12}, extended for RMRC 2013 Indoor Segmentation Challenge, where currently the proposed model ranks first, with an average score of 24,61\% and a number of 39 classes won. Moreover, we show that by combining second-order and deep learning features, over 15\% relative accuracy improvements can be additionally achieved. In a scene classification benchmark, our methodology further improves the state of the art by 24\%.

\end{abstract}

\section{Introduction and Related Work}

The problem of semantic segmentation in monocular images is of central importance in areas like robotics, human-computer interaction and scene understanding for large-scale indexing. For intensity images, significant progress has been achieved recently through work performed in association with the VOC Semantic Segmentation challenges\cite{pascal-voc-2012}, where high performing methods for boundary detection\cite{globalPB08,leordeanu_eccv12}, feature description and non-linear feature maps\cite{carreira2012semantic,ren2012rgb,improv_fisher,Vedaldi2010,li_cvpr12,verbeek2013}, image segmentation \cite{ArbelaezMFM11,carreira_pami12,endres_proposal_eccv2010,levinshtein10} as well as optimization and contextual reasoning \cite{torr_enforcing,Ramalingam_08_exact_multilabel,kolhi_robust_08,harmony_10,Ion2011,dann2012pottics,xia_wei_eccv12} have been developed. Recently the use of deep feature extraction learning frameworks, trained on large-scale databases like Imagenet, has been shown to be effective not only for image classification \cite{krizhevsky2012imagenet}, but also for semantic segmentation\cite{girshick2013rich_cvpr}, where in conjunction with figure-ground proposal generation methods\cite{carreira_pami12}, impressive results have been achieved (the effectiveness of such regions descriptors will be analyzed in our proposed RGB-D framework, as well).

The scientific problem of three dimensional scene understanding from images, both quantitative\cite{Zisserman1995239,Forsyth91} and qualitative\cite{Hoiem_2007_5818,saxena_ijcv}, has a long standing research tradition in computer vision. Some of the more recent work has focused on the analysis of cluttered indoor scenes \cite{Hoiem_2007,Lee2010,Gupta2010,Gupta2011,Hedau2012}. In this setup \cite{Lee2010,Hedau2012} analyze the geometry of the rooms including surfaces and objects, whereas \cite{Gupta2011} reason about object functionality from the standpoint of a human user of the environment.

The existence of affordable and increasingly miniaturized time of flight and infra-red sensors like Kinect opens the possibility that RGB-D sensors will be embedded in any device, mobile or not in the near future. This creates scientific and technological opportunities for exploiting the RGB-D information for scene understanding and semantic segmentation, with potentially high gains in tasks that have been traditionally considered very challenging when performed based on intensity images alone. Range data has been extensively studied in the past, not only at the level of adapted descriptors like spin images\cite{johnson1999using} and 3D shape contexts\cite{Frome2004} but also for shape modeling using, e.g., deformable superquadrics\cite{Leonardis1997}. 

Besides the recent success for real-time human pose estimation\cite{shotton2013real}, Kinect has also spurred a wave of scene understanding research in robotics\cite{ren2012rgb,tang2012} and computer vision\cite{Silberman:ECCV12,ren2012rgb,gupta2013perceptual,Lin2013holistic,BanicaICCVwks13,Banica2013arxiv_v1} with datasets\cite{silberman2011indoor,Lai2011} recently made available. Our work relates to these recent RGB-D analysis approaches, and we will review them showing how we differentiate in methodology and focus. The NYU Depth V2 dataset was introduced in \cite{Silberman:ECCV12}, where the authors develop an expressive methodology for semantic segmentation by labeling merged superpixels while also inferring support relations between objects. Baseline approaches for semantic segmentation were proposed in \cite{silberman2011indoor}, where multiple alternatives were considered for the unary and pairwise terms inside a pixel-level CRF, with unary terms combining the output of a neural network applied on local descriptors and a depth-sensitive location prior; pairwise terms enforced smoothness while preserving depth discontinuities.
In \cite{ren2012rgb} a superpixel hierarchy is used, and the leaf superpixels are described using concatenated features (kernel descriptors) extracted from the entire path towards the root node of the segmentation tree. 
The recent work in \cite{gupta2013perceptual} achieves excellent results for semantic segmentation after revisiting related problems such as boundary detection, bottom-up grouping and scene classification and extending the methodology to take advantage of depth information. The authors start with a hierarchy of non-overlapping (superpixel) partitions, and use the long-range amodal completion of surfaces for better region grouping. 

Our methodology differentiates from the above approaches in our multiple figure-ground proposal generation based on parametric max-flow extended to use intensity and depth information\footnote{Note that in parallel with the initial versions of our work \cite{BanicaICCVwks13,Banica2013arxiv_v1}, ideas based on our earlier RGB-based constrained parametric min cuts (CPMC) \cite{carreira_pami12} and second-order pooling (O2P) \cite{carreira2012semantic} have also been used for RGB-D data in \cite{Lin2013holistic}. In any case, notice however, that \cite{Lin2013holistic} address the different task of 3D object detection, providing methodology to assign labels to 3D cuboids, instead of a pixel-level segmentation, as our focus in this work.}, as well as in the feature description, second order pooling, and inference procedure used, which is adapted to handle RGB-D models with many categories and scenes where many instances are present, at widely varying spatial scales. Our pooling process operates over descriptors that capture both appearance (e.g. SIFT \cite{lowe2004distinctive}) and geometry (e.g. spin images \cite{johnson1999using}). Besides the pooled local descriptors we also extracted point cloud features to coarsely characterize the aspect and size of each region. Also, we extracted features from a large convolutional neural network trained for image classification on ImageNet. In the experiments we report the performance of the above features individually and also show the benefits of using them together to jointly describe each region. A class label is assigned to each segment by learning linear category models, one per class, trained to predict the overlap (IoU) between the segment and the best matching object of that class. Finally, an inference procedure is defined in order to resolve conflicts 
between overlapping segments which were assigned different labels, and to generate a final per-pixel segmentation. We analyze the effectiveness of integrating depth, as well as the proposed solutions at each stage of this pipeline, perform analysis of alternative features including those obtained from deep learning, and show that in the challenging NY Depth v2 dataset \cite{Silberman:ECCV12}, extended for RMRC 2013 Indoor Segmentation Challenge, the proposed model ranks first, with an average score of 24,61\% and a number of 39 classes won.

The rest of the paper is organized as follows: sec. \ref{sec:CPMC-3D} presents how depth data is used in order to improve the generation of figure-ground segmentations within parametric max-flow models, sec. \ref{sec:labeling} illustrates the procedure used for assigning a label to each segment, while sec. \ref{sec:inference} describes the procedure we used for resolving the conflicts between overlapping segments in order to obtain the final per-pixel labels. Experiments follow in sec. \ref{sec:experiments}. We conclude and discuss ideas for future work in sec. \ref{sec:conclusions}.

\section{Parametric Generation of Figure-Ground Proposals}
\label{sec:CPMC-3D}

In contrast to methodologies that compute hierarchical, non-overlapping partitions of the image into multiple regions, our approach relies on generating multiple overlapping figure-ground segmentations, systematically, based on parametric max-flow solvers. We focus on constrained parametric min cuts models CPMC\cite{carreira_pami12} generalized to take advantage of intensity and depth information (CPMC-3D). We rely on simple spatial energy models based on attention mechanisms that allow us to solve for all breakpoints (segmentation solutions), corresponding to different locations and spatial scales, in polynomial time. The idea is to `fixate' at different spatial locations, set up constraints such that a fixated location is assigned to the foreground, and elements on the boundary of the image are assigned to the background, then solve for the set of binary partitions that can be obtained under such constraints. Because solutions obtained at different fixation points may overlap, or may have low quality, skewed shape statistics, a ranking process ensures that only a valid and compact subset is retained. The ranker (in our case a linear regressor) is trained to distinguish between those segments that exhibit the regularities of real-world objects (e.g. continuity, convexity, Euler structure, etc.) and the ones that do not. This `objectness' criteria is category independent: the ranker is trained using a large variety of shapes belonging to many visual categories. Following duplicate elimination and hypothesis scoring, a Maximal Marginal diversification stage ensures that the pool of solutions obtained contains good quality configurations that are sufficiently different from each other.

The figure-ground segmentation proposals are generated by solving a family of optimization problems for spatial energies of the form:
\begin{equation}
E^\lambda(L) =\sum_{x} D_\lambda(l_x) + \sum_{x, y \in \N(x)} V_{xy}(l_x, l_y)
\label{eq:cpmc-energy}
\end{equation}
where $L$ is a labeling of the pixels in the image into foreground or background, $\N(x)$ is the neighborhood of a particular pixel/node $x$, $\lambda\in\mathbb{R}$ selects the problem instance to be solved, the unary term $D_\lambda$ defines the cost of assigning a particular pixel to the foreground or the background, and the pairwise term $V_{xy}$ penalizes the assignment of different labels to similar neighboring pixels.
\vspace{-3mm}
\begin{figure}
 \begin{center}
  \includegraphics[width=0.90\linewidth]{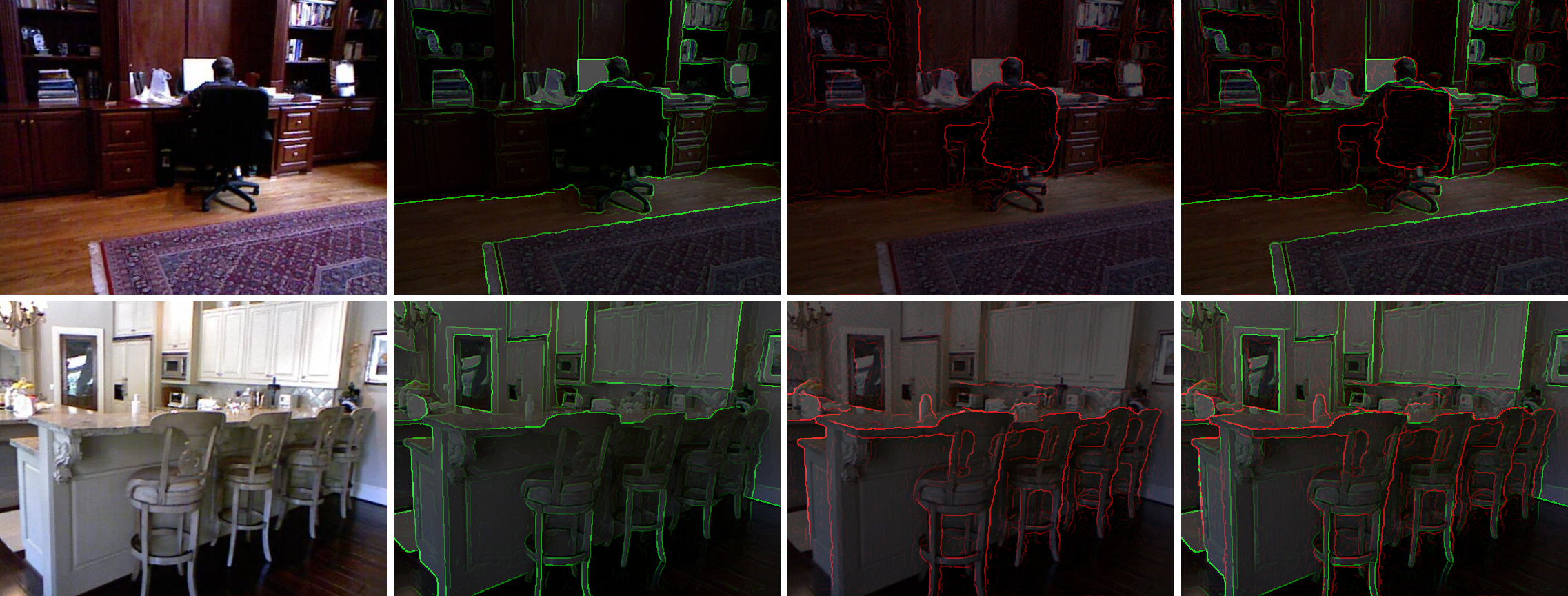}
 \end{center}
\vspace{-5mm}
 \caption{
  Depth and intensity boundary complement each other. Left to right: original image, boundaries extracted from the intensity image, boundaries extracted from the depth image, boundaries resulting from the fusion of RGB and depth information, {\it cf} \eqref{eq:boundaries} with winning channel shown.
 }
 \label{fig:boundaries-rgbd}
\end{figure}
\vspace{-5mm}

\begin{figure}[!ht]
	\begin{center}
		\includegraphics[width=0.85\linewidth]{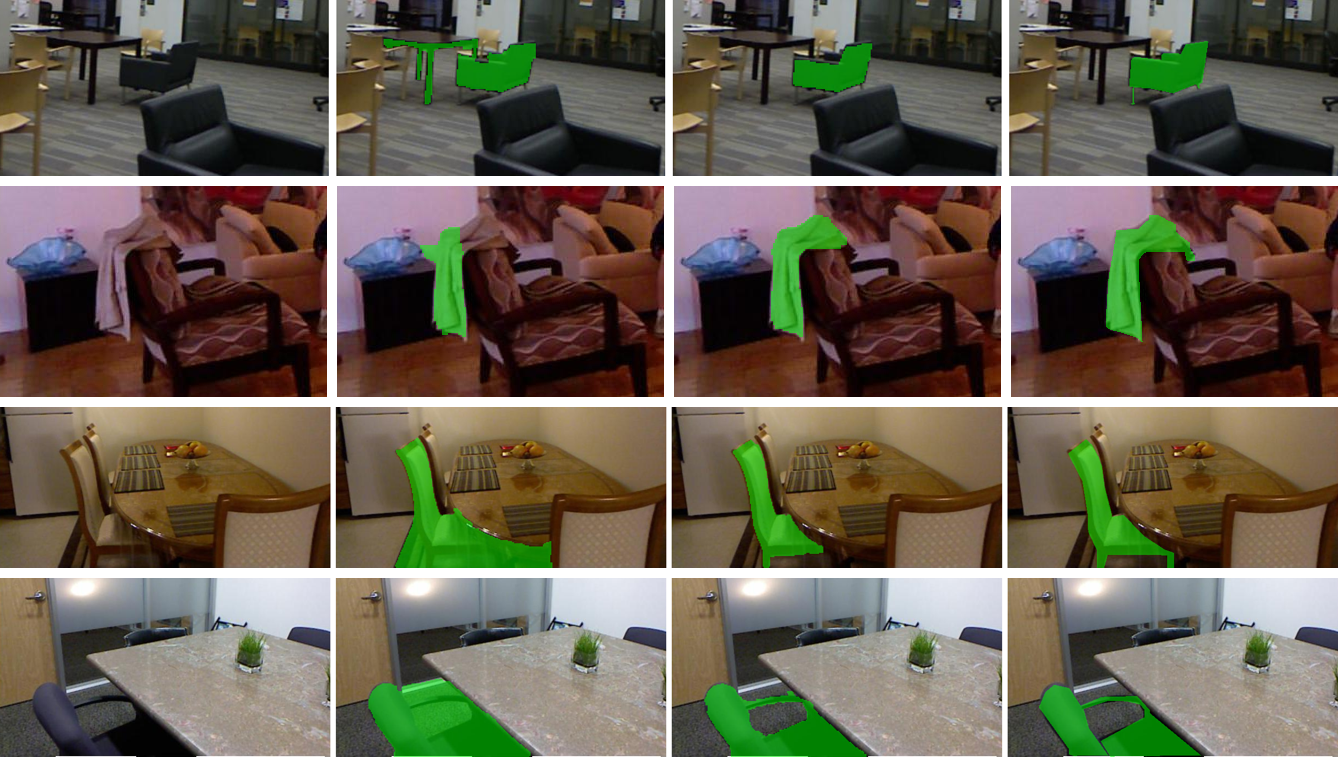}
	\end{center}
\vspace{-5mm}
	\caption{
		By combining depth and intensity cues we can significantly improve the quality of the figure-ground proposal. Left to right: original image, best segment obtained from constrained parametric max-flow on intensity images (CPMC), best segment from CPMC-3D that combines intensity and depth information, and ground truth. The images are from the NYU Depth V2 dataset \cite{Silberman:ECCV12}.
	}
	\label{fig:cpmc-rgbd1}
\end{figure}

In order to incorporate depth data, our CPMC-3D model relies on two boundary probability maps, one based on the RGB information in the image and the other using the depth image. We then modulate the pairwise term of the spatial model to account for both intensity and depth discontinuities, resulting in a more accurate pool of segments (see \fig{fig:cpmc-rgbd1} for qualitative results). The intensity-based pairwise term $V_{xy}$ in eq.~\ref{eq:cpmc-energy} has the following form: $g(x,y) = \exp\left[-\frac{\max(Gb_\I(x), Gb_\I(y))}{\sigma^2} \right]$ when two neighboring pixels $x,y$ are assigned different labels, where $Gb_\I$ is the output of a generalized, trained contour detector~\cite{leordeanu_eccv12,globalPB08} computed for the image $\I$ at a given pixel. In order to fuse depth information, we define an augmented penalty:
\begin{equation}
g'(x,y) = \exp\left[-\frac{\max(Gb_\I(x), Gb_\I(y), Gb_{\D}(x), Gb_{\D}(y))}{\sigma^2} \right]
\end{equation}\label{eq:boundaries}
where $Gb_{\D}$ is the output of a global contour detector \cite{leordeanu_eccv12,globalPB08} on the depth image. The effects of the proposed boundary fusion scheme are illustrated in \fig{fig:boundaries-rgbd} where it can be seen that we can adaptively select useful boundaries using both RGB and depth cues.

By solving for $\min_{\lambda,L} E^\lambda(L)$ of the sub-modular energy using parametric max-flow, we systematically obtain an entire family of nested solutions in polynomial time (the nesting property of the solutions for this model enables an efficient solver for all breakpoints). For imaging models, the nesting property also ensures that solutions are obtained at different spatial scales in the image -- provided that our `attention mechanism' operates over a sufficiently fine grid, both small and large objects are usually covered quite well. The segments are ranked using a class independent predictor, based on the object-like regularities that each region exposes. In our experiments, we use this ranking to retain, only the top $K=500$ scoring hypotheses for further processing.

Fig. \ref{fig:cpmc-rgbd} illustrates how better segment pools are obtained by fusing RGB and depth information in CPMC-3D. Notice that thin structures (considering the detail available at that spatial scale) and fine details of objects are captured extremely well -- see for instance the legs or the arm rest of chairs. This is promising for robotic RGB-D sensing systems that would be capable to both recognize and manipulate objects in the long run. Quantitatively, the improvement due to the usage of depth is also significant (\S\ref{sec:experiments}).

\section{Description and Recognition of Regions} 
\label{sec:labeling}

Our RGB-D semantic segmentation model relies on the individual description and confidence level recognition of region proposals, that are later fused in  order to construct a complete scene interpretation where conflicts in overlapping partitions are resolved, so that each pixel in the image is accounted for (i.e. assigned a label).

To assign an object class label to each of the $K$ retained region proposals, we train multiple models, one for each class, to predict the overlap (IoU) of a given region with the best-matching ground truth object of that class. At test time, we evaluate all predictors (in our case linear support vector regressors) for each region and assign the class label to the category model which predicts the highest overlap.

The segment description passed to the category-level predictive models is a non-linear feature vector consisting of two components. First, we densely extract local descriptors inside the free-form region and aggregate them using second order pooling O2P\cite{carreira2012semantic}. Another component of the feature vector is obtained by extracting size and aspect statistics from the 3D bounding box that encloses the point cloud corresponding to the region. Both of these, together with  the training and testing of category predictors are described in the next subsections.

\subsection{Second-Order Pooling Over Local RGB-D Descriptors}
\label{sec:o2p}
To characterize a proposal region, we use local descriptors that capture both the appearance and the depth information available in the RGB-D images. Local descriptors extracted inside the region are aggregated using Second Order Pooling (O2P) \cite{carreira2012semantic}. O2P introduces multiplicative second-order analogues of average pooling that together with additional non-linearities (matrix logarithm, power normalization) produce good predictors without the need of going through a feature coding step.

We pool local features characterized by say, $M$ descriptors, $X=(\textbf{x}_1, \ldots, \textbf{x}_M)$, $\textbf{x} \in \mathbb{R}^n$, extracted over patches centered at image locations sampled inside the region proposal $R$, to form global descriptors based on second-order statistics. We will exploit multiplicative second-order interactions (e.g. outer products), with average operators. We define \textit{second-order average-pooling} (2AvgP) as the matrix:

\begin{equation}
\label{2nd_order_avg_pooling}
\textbf{G}_{avg}(R) = \frac{1}{M} \sum_{i} \textbf{x}_i \cdot \textbf{x}_{i}^{\top},
\end{equation}

As the second order pooling operator construct a symmetric positive definite matrix, we will use the log-Euclidean metric adapted for this space. The Log-Euclidean framework employs the simple matrix logarithm as mapping, resulting in a space of $\mbx{S}_{n}^+$ that is \tx{isomorphic} (the algebraic structure of the vector space is conserved), \tx{diffeomorphic} and \tx{isometric} (distances are conserved) to the associated Euclidean space of symmetric matrices. The matrix logarithm can be viewed as the logarithm map with base point set at the identity matrix $\mbx{I}_n$. We apply this operator on the second-order statistics $\textbf{G}_{avg}$ of each region proposal $R_j$, generated using CPMC-3D: 

\begin{equation}
\textbf{G}_{avg}^{log}(R) \leftarrow \log \left( \textbf{G}_{avg}(R) \right),
\end{equation}

The logarithm is obtained using the Schur-Parlett algorithm which takes $\mathcal{O}(n^3 \div n^4)$  operations depending on the distribution of eigenvalues of the input matrices.

Our pooling process considers both RGB and depth information. We first pool features that have proven effective for RGB data \cite{carreira2012semantic} -- SIFT, masked SIFT, Local Binary Patterns (the LBP descriptor). In order to exploit the additional depth information available, we pool over spin images\cite{johnson1999using}, masked spin images and SIFT, masked SIFT, and Local Binary Patterns applied to the depth image. The main differences between the masked and non-masked version of a descriptor occur at those points near the boundaries of the region, where the spatial support of the local descriptor may include fragments outside the current region, belonging to other objects -- choosing to ignore the points outside the current region leads to the masked version of the descriptor. The 3D local descriptors are further enriched using location and color information.

\subsection{Structural 3D Point Cloud Features}
In order to better characterize the structure of a region proposal, we additionally extract a series of measurements from the 3D bounding box of the point cloud associated to it. We characterize the 3D bounding box of the region proposal by 11 numbers: volume, surface, diagonal, perimeter (sum of all side lengths), min side length, median side length, max side length, the length of each side along the 3 axes, and aspect ratio (min side / max side). 
Fitting a bounding parallelepiped to a region point cloud exactly may not produce desirable results due to noise. Therefore in order to achieve robustness we ignore a fixed percent extremal points along each of the 3 axes. This outlier percent was varied (0\%, 2.5\%, 5\%, 7.5\%), to generate the 11-dimensional feature vector for each threshold process. 
We combined the 4 levels to obtain a 44 dimensional descriptor for the point cloud, then let the classifier decide what represents a good threshold.

\subsection{Confidence Models for Region Categories}

The second order RGB-D descriptors and the 3D point cloud features described in the previous two sections are concatenated and used as a joint region descriptor. For each category we train linear regression models to predict the overlap between a region and the best-matching objects of each class -- one predictive model is trained for each category.
The data used for building the predictive category models is composed of the features extracted on the ground truth masks from the training set along with the $K$ masks generated by CPMC-3D for each training image, with their true IoU overlap with the ground truth. For the ground truth masks the target value will be 1 for the predictive model associated to the specific class of the object, and 0 for all other models, whereas for the imperfect CPMC-3D segments the target output will be a value in the $[0, 1]$ interval.

At test time, we assign a class label to each of the $K$ retained masks by running all category predictors and choosing the class with maximal estimated overlap. The regression model naturally provides a useful confidence measure, for each proposal and visual category. While this provides a decision at the level of regions considered in isolation, such regions may overlap. In order to construct the final solution, the predicted labels of regions together with their confidence will be used within a sequential inference process that resolves conflicts and assigns labels for the entire image.

\section{Sequential Inference with Confidence Reasoning}
\label{sec:inference}

At this point, for a given test image, we have $K$ overlapping object-level proposals which have been independently labeled to visual categories using the methodology just described. We also have confidences for estimates. Our objective is to generate a per-pixel labeling. This requires a procedure to resolve the conflicts between overlapping segments with different labels. We begin by further reducing the segment pool, by retaining only the most confident $S$ regions based on the class-specific outputs from category predictors. 

We sequentially analyze each region and let it compete with previous overlapping segments that have already been included in the labeling of the image. Specifically, at any step of the sequential procedure we have a partial annotation of the image obtained from the segments considered so far, along with information regarding the segment chosen to label each particular pixel. Then a segment (referred to as the `current segment') attempts to re-label portions of the image. To decide where and whether to re-label or not, we independently resolve the competitions between the current segment and each previous segment that is part of the current solution, and partially overlaps the spatial support of the current one. We resolve conflicts one segment at a time. We denote the `conflict region' the area inside the spatial extent of the current segment that was labeled differently by a previous segment and analyze criteria to decide whether to assign the conflict region to the current segment or maintain its original label. Two conflict resolution strategies have been considered:

\noindent{\it Overlap:} A simple yet effective procedure proved to be the one of always assigning the conflict region to the segment that has larger overlap (IoU) with it. This translates into selecting the segment with the smaller area, since the conflict region is included in both segments\footnote{ Note that it may be implemented efficiently by sequentially placing segments on top of each other, with the larger ones added first, allowing the later smaller ones to overwrite their labels where overlaps occur. This reduces to computing the size of each mask ($\mathcal{O}(W \cdot H \cdot K)$ complexity) ordering them by size -- $\mathcal{O}(K \cdot \log(K))$ -- and then making a final pass through all the sorted masks in reverse order of size, which results in an overall $\mathcal{O}(W \cdot H \cdot K)$ complexity per image, since $K$, the number of masks, is much smaller than the number of pixels in the image -- this takes below 0.5 seconds per image in our Matlab implementation, for $W=640$, $H=480$, $K=500$.}. 

\noindent{\it Overlap and Confidence:} We also tested criteria based on the confidence provided by the category-specific predictors. A straightforward variant always assigns the conflict region to the segment with higher confidence regarding its label. In this case performance decreased, as the procedure is prone to falling into the trap of letting a single large segment label most of the image.
To palliate this effect we experimented with a procedure that combines confidence (the category specific predictor response) and the proposal size. In the implementation we use the confidence output of the predictor only when the difference in size is not dramatic: if one segment's size is smaller than 50\% of the other's we use the small one, otherwise (they have similar size) we decide based on confidence. In our experiments, this combination between size and confidence achieved better results than each one of them considered in isolation.


\section{Experiments}\label{sec:experiments}
Our experiments were conducted on the NYU Depth V2 dataset \cite{Silberman:ECCV12}, which contains 1449 RGB-D images. We model 92 object classes for semantic labeling, each being found at least 50 times in the NYU Depth V2 dataset. The object classes are those used for the RMRC Indoor Segmentation Challenge at ICCV 2013. We also report scores obtained after evaluating on the test set on the dataset server.

In our implementation the spin images pooled using O2P were represented by $16\times16$ 2D histograms, extracted at two spatial scales -- considering points within a radius of 0.3 respectively 0.5 meters. The RGB-based local descriptors (SIFT and LBP) were computed using the same parameters as in the publicly available implementation of O2P. In the PCA reduction step we retained 2,500 dimensions from the pooled spin images and 2,500 dimensions from the pooled masked spin images, along with the 12,500 dimensions retained from the descriptors which use RGB information (SIFT and LBP). When pooling SIFT, masked SIFTs and LBP descriptors on the depth image we used the same parameters as for RGB, but when reducing the dimensionality of the descriptors using PCA we retained 2,500 dimensions from each descriptor type (instead of retaining 5,000 dimensions as in RGB for each variant of the SIFT descriptors -- masked/not masked).

We have also experimented with `deep features' extracted from a large convolutional neural network trained for image classification on ImageNet. We followed the procedure and implementation from \cite{girshick2013rich_cvpr}, without the fine-tunning step, which resulted in a 4096 dimensional feature vector.

We next analyze the effects of various components of the system, at each stage. Unless otherwise indicated, the results reported below are obtained on the test set of NYU Depth V2, using the standard train-test split which consists of 795 training images and 654 testing images.

\begin{table}[tbpH]
\centering
\begin{tabular} {|c|c|c|c|}
	\hline
	CPMC	&	CPMC-3D	&	Combined	&	Upper bound\\
	\hline
 	0.57  	& 	0.59	&	0.62	&	0.68\\
	\hline
\end{tabular} 
\caption{
	Integrating RGB and depth cues generates improved figure-ground segmentations. The values represent the average IoU measures over ground-truth objects for the best-matching proposal and indicate the scores for the CPMC algorithm (\cite{carreira_pami12}), the CPMC-3D algorithm (sec. \ref{sec:CPMC-3D}), the score obtained when using the union of the pools generated by CPMC and CPMC-3D, and an upper bound. The upper bound is generated assuming that perfect (gt) boundaries are available.
}
\label{table:cpmc-3d}
\end{table}

\begin{figure}
	\begin{center}
		\includegraphics[width=1.0\linewidth]{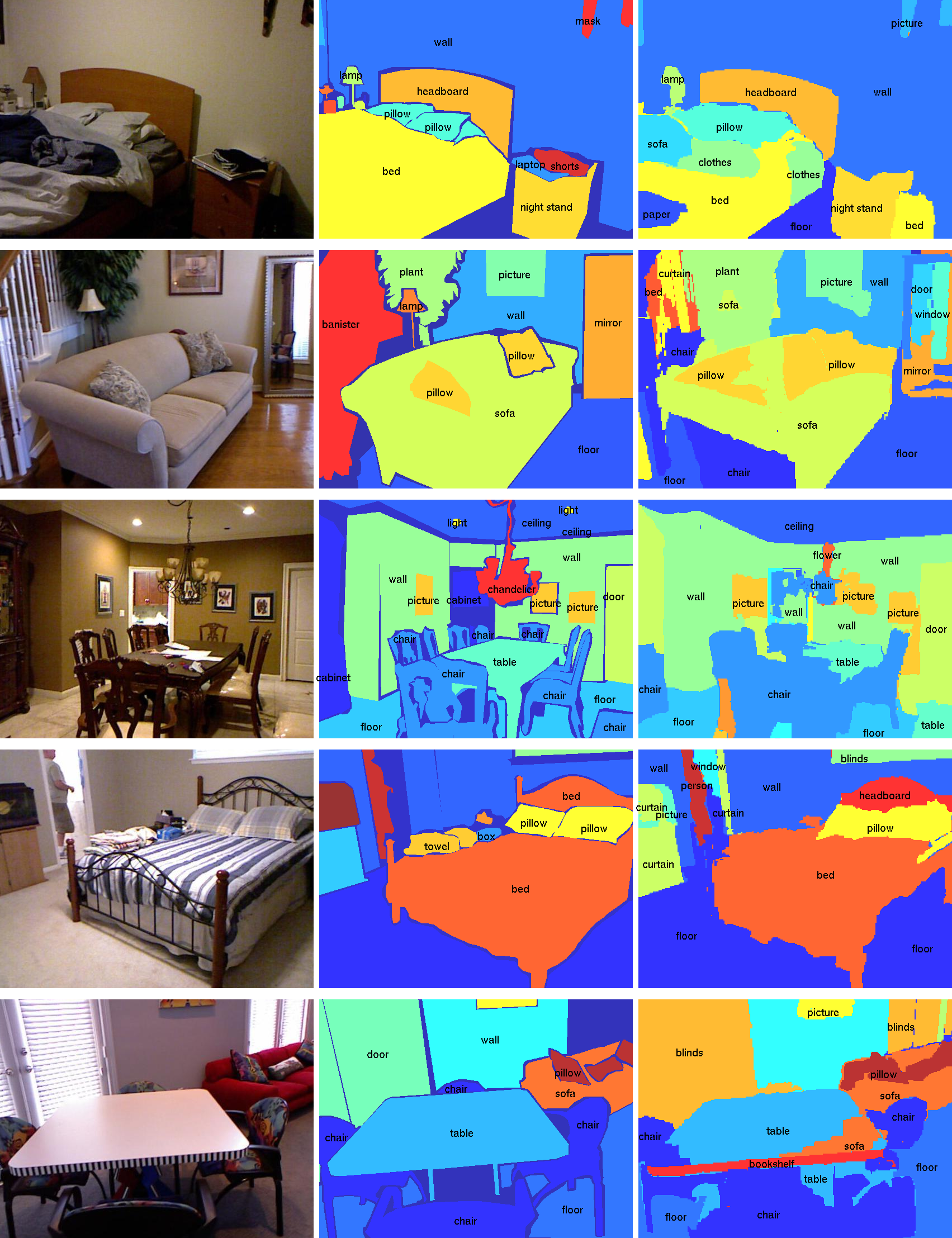}
	\end{center}
\vspace{-5mm}
	\caption{
		Sample semantic segmentations generated by our system. Left to right: RGB image, ground truth semantic segmentation, our segmentation.
	}
	\label{fig:cpmc-rgbd}
\end{figure}

\noindent{\bf Parametric Generation of Figure-Ground Proposals:} We have generated proposals using a regular `attention model' based on a 5x5 grid of seeds, and constraints placed as described in sec. \ref{sec:CPMC-3D}. We first investigated the impact of depth in the generation of the segment pool. We show qualitative results in fig. \ref{fig:cpmc-rgbd} and quantitative ones in table \ref{table:cpmc-3d}. 

\noindent{\bf Description and Recognition of Regions:} After extracting multiple figure-ground segment proposals based on RGB-D, each of them is categorized, with confidence, using the procedure described in sec. \ref{sec:labeling}. We retained $K=500$ segments from each testing image (the highest-scoring regions according to a category-independent ranker). For training we used both the clean ground truth masks and noisier automatically generated segment proposals. We observed only marginal improvements when training with more than 300 masks per image -- therefore we only retained 300 segments for training, which are passed to category-specific predictors, along with ground truth segments. Notice that we use 300 proposals in training and 500 in testing. There is no inconsistency as these numbers need not be the same -- in practice we have also experimented with mixed regular and irregular grids where we made sure that we always placed seeds on ground truth objects in training, but this strategy did not produce significantly better pools than the ones based on a regular 5x5 seeding grid.

We will extensively analyze the performance of the segment descriptors constructed based on both RGB and depth information. We report intermediary results as well since the inference process that estimates per-pixel segmentations involves steps which are in turn prone to error. 

\noindent{\bf Labeling Ground Truth Segments:} We begin by analyzing the performance of our descriptors on the clean ground truth segments from the NYU Depth V2 test set. Results are shown in table \ref{table:label-gt-masks}. Interestingly, the pooled depth descriptors performed better than the RGB descriptors. However, their combination significantly boosted the score, confirming that indeed complementary information is present in the depth and intensity channels, and our model can leverage it.

\begin{table}[tbpH]
\centering
\begin{tabular} {|c|c|c|c|c|c|c|c|c|c|}
\hline

\multirow{3}{*}{\begin{tabular}{@{}c@{}}Deep \\ features\end{tabular}}		 
				&	\multicolumn{6}{|c|}{O2P on local descriptors}															&	\multirow{3}{*}{PCF}	&	\multirow{3}{*}{\begin{tabular}{@{}c@{}}O2P \\ + \\ PCF \end{tabular}}	&	\multirow{3}{*}{\begin{tabular}{@{}c@{}}O2P \\ + \\ deep feats.\end{tabular}}	\\
\cline{2-7}
				&	 all		&	\multicolumn{4}{c|}{Depth features}					&		\multirow{2}{*}{\begin{tabular}{@{}c@{}}all \\ RGB-D\end{tabular}}	&							&							&	\\
\cline{3-6}
				&	 RGB	&	spin imgs  & SIFT depth & LBP depth & all depth	&									&							&							&	\\
\hline
	45.43		&	55.98		&	47.04	   & 52.39		& 40.84		& 57.22			&			62.95					& 			16.46 			& 			62.94			&	64.54	\\
	
\hline

\end{tabular} 
\caption{
	Accuracy of different RGB and depth descriptors in labeling the ground-truth segments on the NYU Depth V2 test set.
}
\label{table:label-gt-masks}
\end{table}


\noindent\textbf{Labeling Figure-Ground RGB-D Segment Proposals:} We next analyzed the behavior of the descriptors considering the segments generated by our parametric solver operating on RGB-D channels. This aims to analyze the robustness of the descriptors with respect to imperfections in segmentation. Categorizing segments individually is the final step before proceeding to the inference described in sec. \ref{sec:inference} where overlapping segments compete for pixel labeling. The performance of labeling imperfect segments is given in table \ref{table:label-CPMC-masks}.

\begin{table}[tbpH]
\centering
\begin{tabular} {|c|c|c|c|c|c|c|c|c|c|}
\hline

\multirow{3}{*}{\begin{tabular}{@{}c@{}}Deep \\ features\end{tabular}}		 
				&	\multicolumn{6}{|c|}{O2P on local descriptors}															&	\multirow{3}{*}{PCF}	&	\multirow{3}{*}{\begin{tabular}{@{}c@{}}O2P \\ + \\ PCF \end{tabular}}	&	\multirow{3}{*}{\begin{tabular}{@{}c@{}}O2P \\ + \\ deep feats.\end{tabular}}	\\
\cline{2-7}
				&	 all		&	\multicolumn{4}{c|}{Depth features}					&		\multirow{2}{*}{\begin{tabular}{@{}c@{}}all \\ RGB-D\end{tabular}}	&							&							&	\\
\cline{3-6}
				&	 RGB	&	spin imgs  & SIFT depth & LBP depth & all depth	&									&							&							&	\\
\hline
61.69			&	56.87		&	46.01	   & 54.90		& 46.27		& 59.35			&			65.22					& 			12.87 			& 			65.54	&	67.15\\
\hline

\end{tabular} 
\caption{
	Accuracy for labeling figure-ground RGB-D proposals extracted automatically, on the NYU Depth V2 test set. The correct label of a proposal is assumed to be the label of the ground truth object that mostly overlaps that segment. Only segments that have at least 50\% overlap with a ground truth object are considered.
}
\label{table:label-CPMC-masks}
\end{table}

\noindent\textbf{Semantic Segmentation:} 
In table \ref{table:per-pixel-labeling}, we report the end-to-end performance using various descriptors for labeling segments. The metric is the one used in the RMRC Indoor Semantic Segmentation Challenge held during ICCV 2013 -- mean recall per class. In table \ref{table:final-results} we show the scores of our segmentations, which were uploaded on March 06, 2014 (original date of this paper revision) on the RMRC test server. These segmentations were generated using only the pooled local descriptors (O2P).

\begin{table}[tbpH]
\centering
\begin{tabular} {|c|c|c|c|c|c|c|c|c|c|}
\hline

\multirow{3}{*}{\begin{tabular}{@{}c@{}}Deep \\ features\end{tabular}}		 
				&	\multicolumn{6}{|c|}{O2P on local descriptors}															&	\multirow{3}{*}{PCF}	&	\multirow{3}{*}{\begin{tabular}{@{}c@{}}O2P \\ + \\ PCF \end{tabular}}	&	\multirow{3}{*}{\begin{tabular}{@{}c@{}}O2P \\ + \\ deep feats.\end{tabular}}	\\
\cline{2-7}
				&	 all		&	\multicolumn{4}{c|}{Depth features}					&		\multirow{2}{*}{\begin{tabular}{@{}c@{}}all \\ RGB-D\end{tabular}}	&							&							&	\\
\cline{3-6}
				&	 RGB	&	spin imgs  & SIFT depth & LBP depth & all depth	&									&							&							&	\\
\hline
	20.80		&	18.68		&	13.13	   & 16.64		& 11.06		& 20.49			&			24.68					& 			3.28 			& 			24.10		&	29.03	\\
\hline

\end{tabular} 
\caption{
	Semantic segmentation performance on the NYU Depth V2 test set under the average recall per class metric.
}
\label{table:per-pixel-labeling}
\end{table}

\begin{table}[tbpH]
\centering
\begin{tabular} {|c|c|c|}
\hline
\textbf{Method} & \textbf{Number of classes won} & \textbf{Average score} \\
\hline
\hline
Gupta et al. \cite{gupta2013perceptual}	& 32 & 23.98 \\
\hline
Silberman et al. \cite{Silberman:ECCV12}	& 29 & 21.31 \\
\hline
Ren et al. \cite{ren2012rgb}	& 22 & 17.52 \\
\hline
Ours	& \textbf{39} & \textbf{24.61} \\
\hline
\end{tabular} 
\caption{
	Semantic segmentation performance under the average recall per class metric, for 92 classes. The reported results are obtained on the RMRC test set (an extension of the NYU Depth V2 dataset) after uploading our results on the evaluation server.
The metric is the average recall per class (`average score' column). We also report the number of classes where each method achieves the highest score (in case of ties, one point is added for each method achieving the highest score).
	The uploaded method uses the O2P descriptors (without deep learning features), and the `overlap + confidence' inference criterion.
}
\label{table:final-results}
\end{table}

After generating segments and predicting labels for each one of them, we generate a per pixel segmentation. However, this is not straightforward since multiple segments with different labels can overlap. In sec. \ref{sec:inference} we described an inference procedure to resolve such conflicts that reduces to considering pairs of conflicting segments independently. Different criteria may be used to resolve these situations, and we quantitatively assess several such choices. Table \ref{table:inference-experiments} shows that relatively good performance is achieved using a simple overlap criterion which always assigns the conflict region to the segment which has the larger overlap (intersection over union) with it (i.e. it picks the smaller segment). The reason this proved effective is that it is not prone to errors occurring when a large segment gets to label most of the image. This would lead to a drop in performance especially when the metric is not normalized with respect to object size, i.e. mislabeling a small object generates the same performance penalty as mislabeling a large one.  Adding confidence reasoning further improves the results. Qualitative segmentation results produced by our best model are also shown in \fig{fig:cpmc-rgbd}.

\begin{table}[tbpH]
\centering
\begin{tabular} {|c|c|c|}
\hline
\textbf{Criterion}					&	\textbf{O2P Score} & \textbf{O2P + Deep Features Score}\\
\hline
\hline
overlap								&	24.68	& 	29.03	\\
\hline
confidence							&	19.33	&	22.41   \\
\hline
overlap + confidence				&	25.26	&	29.71   \\
\hline
ground truth (upper bound)			&	38.07	& 	41.08	\\
\hline
\end{tabular} 
\caption{
	Different criteria for resolving conflicts between overlapping segments with conflicting category labels. The `overlap' criterion always selects the segment with the larger overlap (i.e. smaller size); `overlap + confidence' selects the segment with the larger confidence as long as there is no dramatic difference (less than 50\%) in overlap with the conflicting region. The `ground truth' represents the maximum achievable score when using such an inference procedure -- given the extracted segments and the predicted labels this is obtained by selecting (based on ground-truth) that region correctly classified (if there is one dominantly and consistently labeled with the ground-truth, inside the conflict region).
}
\label{table:inference-experiments}
\end{table}

\newcolumntype{C}[1]{>{\centering\let\newline\\\arraybackslash\hspace{0pt}}m{#1}}
\newcolumntype{D}{>{\centering\let\newline\\\arraybackslash\hspace{0pt}}m{1.0cm}}

\begin{table}[tbpH]
\centering
\scalebox{0.8}{
\begin{tabular} {|D|D|D|D|D|D|D|D|D|D|D|D|D|}
\hline
Method &  alarm clock &  backpack &  bag &  basket &  bathtub &  bed &  bin &  blanket &  blinds &  book &  books &  bookshelf\\
\hline
Ours &  0 &  0 &  5.5 & \textbf{45.1} &  46 &  32.8 &  0 &  0 & \textbf{48.6} & \textbf{5.9} & \textbf{48.5} &  6.7\\
\cite{gupta2013perceptual} & \textbf{6.9} &  4.6 &  4.4 &  10.9 &  64.2 &  53 &  0 &  8.2 &  24.5 &  2.9 &  35.2 &  6.7\\
\cite{Silberman:ECCV12} &  0 & \textbf{38.9} &  2.8 &  10.5 & \textbf{65.5} &  43.6 &  0 & \textbf{11.2} &  43.6 &  2 &  23.8 &  10.4\\
\cite{ren2012rgb} &  0 &  0 & \textbf{6.9} &  13.1 &  14.2 & \textbf{54.8} &  0 &  2.4 &  14.1 &  2 &  17 & \textbf{15.6}\\
\hline
\hline
Method &  bottle &  bowl &  box &  cabinet &  candle &  candlestick &  ceiling &  chair &  clock &  clothes &  coffee machine &  coffee table\\
\hline
Ours &  14.9 &  0 &  6.7 &  29.5 &  0 & \textbf{100} &  33.3 & \textbf{44.9} &  0 & \textbf{30} & \textbf{27.1} &  0\\
\cite{gupta2013perceptual} &  10 &  2.9 &  8.4 &  25.6 &  0 & \textbf{100} &  80.5 &  29.1 &  0 &  22.5 &  26.2 &  0\\
\cite{Silberman:ECCV12} &  11.1 &  0 &  0.3 &  4.7 &  0 & \textbf{100} & \textbf{85.2} &  11.4 &  0 &  6.7 &  3.2 &  0\\
\cite{ren2012rgb} & \textbf{24.8} & \textbf{9.7} & \textbf{11.3} & \textbf{33.3} &  0 & \textbf{100} &  71.7 &  32.8 &  0 &  20.5 &  12.6 &  0\\
\hline
\hline
Method &  column &  computer &  container &  counter &  cup &  curtain &  desk &  dishwasher &  doll &  door &  door knob &  drawer\\
\hline
Ours &  11.1 &  3.6 & \textbf{100} &  25.6 & \textbf{15.8} & \textbf{61.1} &  15.2 &  0 &  0 &  33.8 & \textbf{11.6} &  3.7\\
\cite{gupta2013perceptual} &  22 & \textbf{6.6} & \textbf{100} & \textbf{44.7} &  3.6 &  32.5 & \textbf{15.7} & \textbf{33.6} & \textbf{42.5} & \textbf{42} &  0 &  0\\
\cite{Silberman:ECCV12} & \textbf{23.9} &  1.7 & \textbf{100} &  17.4 &  8.9 &  40 &  5.8 &  16.1 &  0 &  19.1 &  0 &  0\\
\cite{ren2012rgb} &  1.6 &  0.2 & \textbf{100} &  31.6 &  5.4 &  37.9 &  13 &  10.9 &  0 &  24.4 &  0 & \textbf{6.9}\\
\hline
\hline
Method &  dresser &  drying rack &  electrical outlet &  fan &  faucet &  faucet handle &  floor &  floor mat &  flower &  garbage bin &  headboard &  jar\\
\hline
Ours &  27.2 & \textbf{38.9} &  0 &  2.5 &  5.3 &  0 &  70.2 &  28.8 & \textbf{33.5} & \textbf{27} & \textbf{100} &  0\\
\cite{gupta2013perceptual} &  25.4 &  0 & \textbf{6.9} & \textbf{6.2} &  12.7 &  0 &  74.6 & \textbf{52.5} &  20.2 &  16.6 & \textbf{100} & \textbf{10.8}\\
\cite{Silberman:ECCV12} & \textbf{34.5} &  0 &  0 &  1.2 & \textbf{13.9} &  0 & \textbf{79.2} &  41.4 &  23.9 &  20.1 & \textbf{100} &  1.4\\
\cite{ren2012rgb} &  16.3 &  0 &  0 &  0 &  12.9 &  0 &  69.7 &  18.6 &  7.9 &  12.4 & \textbf{100} &  0.8\\
\hline
\hline
Method &  keyboard &  lamp &  light &  magazine &  microwave &  mirror &  monitor &  night stand &  ottoman &  oven &  paper &  paper towel\\
\hline
Ours &  25.6 & \textbf{32.6} &  6.2 &  4.7 &  24.8 &  23.5 & \textbf{31.4} &  20.9 &  0 &  3.8 & \textbf{47.6} &  69.7\\
\cite{gupta2013perceptual} & \textbf{30} &  22.5 &  10.6 & \textbf{13.2} &  13 & \textbf{25.4} &  27.3 & \textbf{32} & \textbf{49.1} &  8.5 &  16.3 &  56\\
\cite{Silberman:ECCV12} &  0 &  14.9 & \textbf{15} &  0 & \textbf{30.8} &  9.8 &  4.2 &  5.2 &  48.3 & \textbf{24.8} &  14.5 & \textbf{71.2}\\
\cite{ren2012rgb} &  0 &  23.3 &  10.3 &  0 &  5.3 &  18 &  10.2 &  7.4 &  0 &  0.9 &  20 &  14.5\\
\hline
\hline
Method &  person &  picture &  pillow &  pipe &  placemat &  plant &  plant pot &  plate &  printer &  fridge &  remote control &  sculpture\\
\hline
Ours & \textbf{100} &  48.6 & \textbf{66.1} &  0 & \textbf{44.4} &  58.3 &  0.1 &  0 & \textbf{29.9} &  12.3 & \textbf{15} &  0\\
\cite{gupta2013perceptual} & \textbf{100} &  34.3 &  52.3 & \textbf{0.8} &  31.2 &  9.4 &  0.4 & \textbf{4} &  26.1 & \textbf{30.1} &  3.1 &  0\\
\cite{Silberman:ECCV12} & \textbf{100} &  30.6 &  58 &  0 &  0 &  59.8 & \textbf{3} &  0 &  11.3 &  26.5 &  0 &  0\\
\cite{ren2012rgb} & \textbf{100} & \textbf{49.4} &  53.5 &  0 &  0 & \textbf{61.3} &  1 &  0.1 &  12.7 &  5.2 &  0.4 &  0\\
\hline
\hline
Method &  shelves &  shoe &  sink &  sofa &  speaker &  stool &  stove &  stuffed animal &  table &  telephone &  television &  tissue box\\
\hline
Ours &  10.1 &  16.3 &  35.1 & \textbf{31.7} &  0 &  0.3 &  18.5 & \textbf{39.8} & \textbf{29.9} & \textbf{8.6} &  49.2 & \textbf{31.3}\\
\cite{gupta2013perceptual} &  10.5 &  18.2 & \textbf{45.8} &  26.3 &  0.3 &  0.1 & \textbf{33.2} &  16 &  14.2 &  0 &  46.1 &  12\\
\cite{Silberman:ECCV12} &  9.3 & \textbf{24.8} &  42.8 &  4.4 & \textbf{11.6} & \textbf{2} &  24.2 &  25 &  2.8 &  4.2 & \textbf{57.1} &  0.8\\
\cite{ren2012rgb} & \textbf{15.9} &  4 &  24.8 &  26.5 &  0.3 &  0 &  10.6 &  3.3 &  16.5 &  0.2 &  28.5 &  0.3\\
\hline
\hline
Method &  toilet &  towel &  toy &  tray &  vase &  wall &  wall decoration &  window & \multicolumn{2}{|c|}{Number of wins} & \multicolumn{2}{|c|}{Average recall}\\
\hline
Ours &  26.9 & \textbf{34.2} &  2.7 &  0 & \textbf{7.1} & \textbf{59.8} & \textbf{35.9} &  20.6 & \multicolumn{2}{|c|}{\textbf{39}} & \multicolumn{2}{|c|}{\textbf{24.61}}\\
\cite{gupta2013perceptual} &  63.2 &  28.4 &  15.1 & \textbf{2.5} &  0.7 &  51.2 &  0 &  31.1 & \multicolumn{2}{|c|}{32} & \multicolumn{2}{|c|}{23.98}\\
\cite{Silberman:ECCV12} & \textbf{70} &  25 & \textbf{23.3} &  0 &  0 &  58.1 &  0 &  23.4 & \multicolumn{2}{|c|}{29} & \multicolumn{2}{|c|}{21.31}\\
\cite{ren2012rgb} &  35.1 &  17.6 &  3.2 &  0 &  1.3 &  59.7 &  10.6 & \textbf{32.4} & \multicolumn{2}{|c|}{22} & \multicolumn{2}{|c|}{17.52}\\
\hline
\end{tabular}
}
\caption{
	Recall (in percents) for each of the 92 classes in the RMRC test set.
}
\label{table:all-classes}
\end{table}
\noindent{\bf Scene classification:} Motivated by the accuracy of the pooled local descriptors we also tackled the problem of scene classification (also studied in \cite{gupta2013perceptual}) and investigated the improvements that resulted by adding depth information. We applied the second-order pooling machinery on top of the same local descriptors presented in section \ref{sec:o2p}, that capture both appearance and depth. The pooling of local descriptors was done in a spatial pyramid, homogeneously (no segmentation proposals) by dividing the entire image in $1$, $2 \times 2$, $4 \times 4$ grids. State of the art results were achieved, as shown in table \ref{table:scene-classification}.

\begin{table}[tbpH]
\centering
\begin{tabular} {|c|c||c|c|c|}
\hline
Class & Gupta et. al \cite{gupta2013perceptual}	&	RGB	&	Depth	& RGB-D\\
\hline
\hline
bedroom		&	79	&	79.06	&	78.01	&	82.72\\
\hline
kitchen		&	74	&	65.09	&	60.38	&	75.47\\
\hline
living room &	47	&	73.83	&	33.64	&	75.70\\
\hline
bathroom	&	67	&	89.66	&	81.03	&	96.55\\
\hline
dining room	&	47	&	96.36	&	50.91	&	96.36\\
\hline
office		&	24	&	63.16	&	13.16	&	71.05\\
\hline
home office	&	8.3	&	70.83	&	0.00	&	62.50\\
\hline
classroom	&	48	&	69.57	&	52.17	&	82.61\\
\hline
bookstore	&	64	&	100.00	&	72.73	&	100.00\\
\hline
others		&	15	&	85.37	&	39.02	&	95.12\\
\hline
mean diag. cm.&	47	&	79.29	&	48.11	&	\textbf{83.81}\\
\hline
avg. accuracy &	58	&	77.52	&	55.81	&	\textbf{82.42}\\
\hline
\end{tabular} 
\caption{
	Scene classification performance on the NYU Depth V2 test set, measured using the mean-diagonal of the normalized confusion matrix (average precision per class) and average classification accuracy. The `RGB' column shows results obtained using descriptors that use RGB data only (SIFT, LBP), pooled using O2P; the `Depth' column gives results using only pooled local descriptors, SIFT, LBP, SPIN computed on depth channels.
}
\label{table:scene-classification}
\end{table}

\section{Conclusions}\label{sec:conclusions}
We have presented a semantic segmentation methodology for RGB-D data, where we have focused on cluttered indoor scenes containing many visual categories. Our approach is based on a parametric figure-ground intensity and depth-constrained proposal process that systematically generates spatial layout hypotheses at multiple locations and scales in the image followed by a sequential inference algorithm that integrates conflicting proposals into a complete scene estimate. We contribute by: (1) generalizing parametric max flow figure-ground methodologies to take advantage of intensity and depth information, (2) region description methods based on second-oder pooling over multiple features constructed using both intensity and depth channels, (3) an inference process that can resolve conflicts in overlapping spatial partitions based on region category confidence reasoning, (4) evaluation of the impact of depth, as well as the effectiveness of a large number of descriptors, both pre-designed and automatically obtained using deep learning, in  a difficult RGB-D semantic segmentation problem with 92 classes. We report state of the art results in the challenging NY Depth v2 dataset \cite{Silberman:ECCV12}, extended for RMRC 2013 Indoor Segmentation Challenge, where the proposed model currently ranks first, with an average score of 24,61\% and a number of 39 classes won. By combining second-order and deep learning features, accuracy improvements in excess of an additional 15\% can be attained. In a RGB-D scene classification benchmark, our methodology further improves the state of the art by 24\%. In future work we plan to study the integration of learned contextual relations in the inference process.
\clearpage\mbox{}

\bibliographystyle{splncs}
\bibliography{rgbd_segmentation}

\end{document}